\documentclass[conference]{IEEEtran}
\ifCLASSINFOpdf
\else
\fi
\usepackage{graphicx} 
\usepackage{multicol}
\usepackage{algorithm}
\usepackage{algpseudocode}
\usepackage{verbatim}
\usepackage{amsmath}
\usepackage{mathtools}

\begin{document}
%
\title{Grasping Unknown Objects in Clutter by Superquadric Representation}

\author{\IEEEauthorblockN{Abhijit Makhal*}
\IEEEauthorblockA{Department of Mechanical Engineering\\
Idaho State University\\
Pocatello, Idaho 83209\\
Email: makhabhi@isu.edu}
\and
\IEEEauthorblockN{Federico Thomas}
\IEEEauthorblockA{
Institut de Rob\`{o}tica i Inform\`{a}tica \\industrial: IRI
(CSIC/UPC)\\
Barcelona, Spain 08028 \\
Email: fthomas@iri.upc.edu}
\and
\IEEEauthorblockN{Alba Perez Gracia}
\IEEEauthorblockA{Department of Mechanical Engineering\\
Idaho State University\\
Pocatello, Idaho 83209\\
Email: perealba@isu.edu}}

\maketitle

\begin{abstract}
In this paper, a quick and efficient method is presented for grasping unknown objects in clutter. The grasping method relies on real-time superquadric (SQ) representation of partial view objects and incomplete object modelling, well suited for unknown symmetric objects in cluttered scenarios which is followed by optimized antipodal grasping. The incomplete object models are processed through a mirroring algorithm that assumes symmetry to first create an approximate complete model and then fit for SQ representation.  

The grasping algorithm is designed for maximum force balance and stability, taking advantage of the quick retrieval of dimension and surface curvature information from the SQ parameters. The pose of the SQs with respect to the direction of gravity is calculated and used together with the parameters of the SQs and specification of the gripper, to select the best direction of approach and contact points. The SQ fitting method has been tested on custom datasets containing objects in isolation as well as in clutter. The grasping algorithm is evaluated on a PR2 and real time results are presented. Initial results indicate that though the method is based on simplistic shape information, it outperforms other learning based grasping algorithms that also work in clutter in terms of time-efficiency and accuracy.
\end{abstract}


%
\IEEEpeerreviewmaketitle

\section{INTRODUCTION}

Accurate grasping of possibly unknown objects is one of the main needs for robotic systems in unstructured environments. In order to do so, robotic systems have to go through multiple costly calculations. First, the object of interest needs to be identified, which includes a segmentation step resulting in a suitable surface impression of the object. In the next stage, the precise position of the object in the world should be determined. Then, grasps needs to be synthesized by calculating the set of points on the object where the fingers can be placed followed by determining preferred approach directions. Finally, the system needs to plan a safe collision-free path of the manipulator towards the object. 

Typical robotic grasping methods focus on optimization of stable grasp metrics. One common approach is to maintain a database of objects with their corresponding preferred grasps. When the system encounters an object that can be identified with an entry from the database, the corresponding suitable referred grasp is pulled out. In this case, the system has to explore and evaluate an increasingly large number of objects and grasps for a given scenario. One of the limitations of such kinds of systems is that the database has to be quite large in order to contain all of the common objects that a robot may encounter. If the system has to work in less structured environments, observing a novel object may be common. A methodology for grasping unknown objects is preferred here, rather than relying on the database. 

\begin{figure}[h!]
\centering
\includegraphics[scale=0.135]{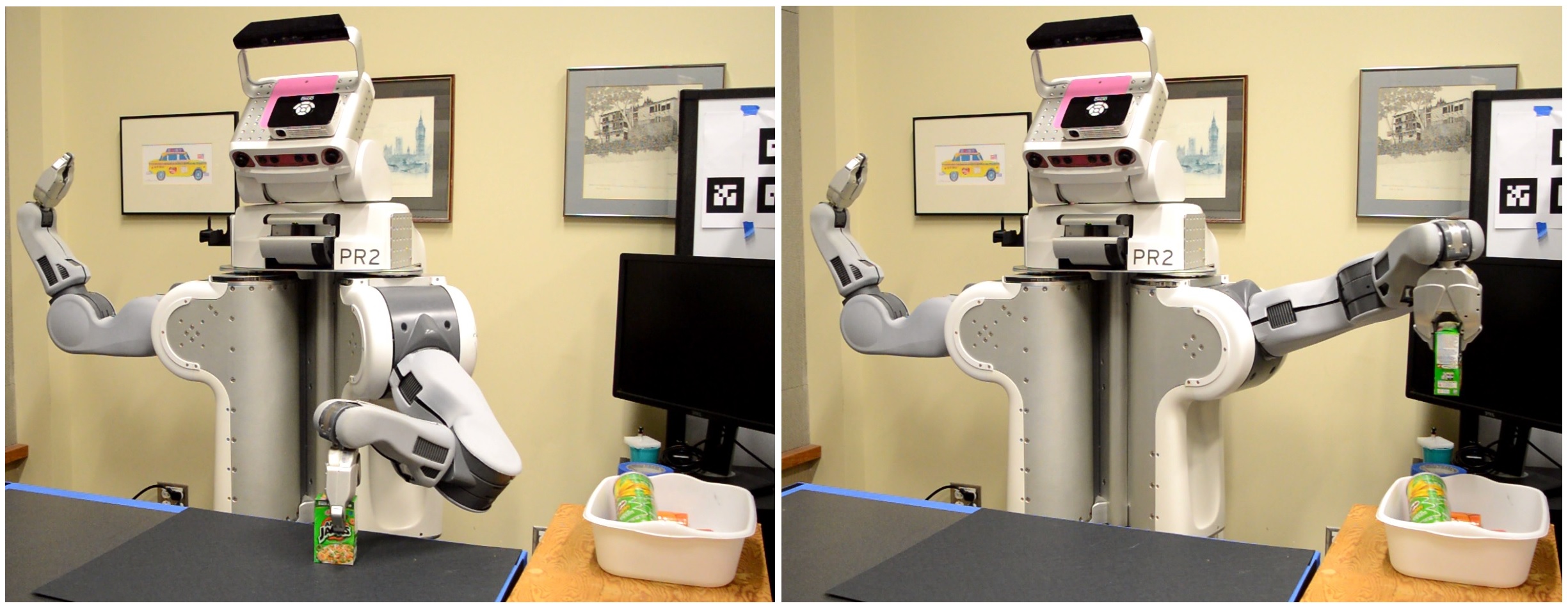}
\caption{Experimental setup.a) The system has to perceive the object on the table, model it by fitting superquadric and execute suitable grasp to pick up the object and b) drop it in the bin to the right of the image}
\label{fig:pr2_setup}
\end{figure}

The rest of the paper is organized as follows: Section \ref{sec:related_work} presents a brief summary of previous work performed for grasping known and unknown objects. Section \ref{sec:superquadric representation} explains the methodology for representation of Superquadrics (SQ). Completion of object models from perceived point cloud is explained in section \ref{sec:model completion}. Section \ref{sec:fitting} describes the process of fitting SQ on completed point clouds, while finding antipodal grasps on those SQ is presented in section \ref{sec:grasp_synth}. The experimental results of fitting SQ on point cloud in real time for objects in isolation and clutter with an evaluation with other methods are presented in section \ref{sec:results}. We conclude with the limitations and bottlenecks of the method in section \ref{sec:conclusion}.

\section{RELATED WORK}\label{sec:related_work}
A considerable amount of work exists in the area of finding feasible grasping points on novel objects using vision based systems. Initial efforts using 2D images to find grasping points in presented in  \cite{Saxena:2008:RGN:1340271.1340280}.  The arrival of economic RGB-D cameras allowed interpreting objects in 3D, which led to object identification as point clouds and derived parameters. Different models have been utilized for this task, such as representing objects by implicit polynomial and algebric invariants \cite{494640},  spherical harmonics \cite{4392735} \cite{kazhdan2003rotation}, geons \cite{wu19933}, generalized cylinders \cite{chuang2004potential},  symmetry-seeking models \cite{Terzopoulos1988} \cite {mgp_approx_symm_sig_06}, blob models \cite{shapiro1980generalized}, union of balls \cite{5653520} and hyperquadrics \cite{924449}, to name a few.

The representation of objects as SQ is found in \cite{1673799} and later in \cite{Jaklic:2000:SRS:358353}. The work in \cite{4209233} shows fast object representation with SQ for household environments. An approach infast and efficient pose recovery of objects using SQ has been presented in \cite{6631176}. In the work of \cite{Katsoulas:2002:FRP:648287.754106} the authors try to fit SQ to piled objects.

Recent literature suggests that using a  single-view point cloud to fit (SQ) can lead to erroneous shape and pose estimation. In general, the 3D sensors are noisy and obtain only a partial view of the object, from a single viewpoint. In order to obtain a full model of the object, several strategies have been developed. In \cite{7139713}, the shape of the object is completed from the partial view and a mesh of the completed model is generated for grasping.  Generating a full model from a partial model can be approached in several ways, mainly symmetry detection \cite{mgp_approx_symm_sig_06} and symmetry plane \cite{1544938} \cite{5650434} \cite{5980354}. Extrusion-based object completion has also been proposed in \cite{6651593}. \cite{6810150} presents a different strategy of changing the viewpoint in a controlled manner and registering all the partial views to create a complete model. This technique provides good results, yet it is not very suitable for real time systems and is prone to errors due to registration of several views. In addition, results are not satisfactory when the working environment becomes densely cluttered.

The calculation of feasible grasping points on a point cloud or a mesh is by nature iterative, hence computationally expensive. In \cite{Platt:2016}, a large set of grasps is generated directly from the point cloud and evaluated using convolutional neural networks, obtaining good grasp success results. Such methods avoid the need for robust segmentation but cannot assure the assignment of the grasp to a target object. The method, denoted Grasp Pose Detection (GPD), has been combined with object pose detection in \cite{Platt:2016b}.  A similar approach uses Height Accumulated Features (HAF) \cite{Vincze:2015},  where local topographical information from the point cloud is retrieved to calculate antipodal grasps.

A different set of methods use the fitting of object models to the point clouds to generate smaller or simpler sets of grasp points. In \cite{Roa:2014}, the calculation of grasp regions is combined with path planning. Curvature-based grasping using antipodal points for differentiable curves, in both convex and concave segments, was studied in \cite{Jia:2002}, while in \cite{Burdick:2002}, a grasping energy function is used to calculate antipodal grasping points using local modelling of the surface.

A third set of methods relies on the recognition of objects and comparison to a database of objects with optimized grasping points already included as features of the object \cite{5980354} \cite{1371616}.

The main contributions of the paper are as follows, 1) a novel method is presented to mirror the partial view point cloud obtaining an approximated full model of the object, 2) SQ fitting on the point cloud as part of the online process which is robust to change in the environment and 3) a novel shape based grasping algorithm which can obtain antipodal grasping points for frictional two-fingered grippers satisfying properties of maximum force balance and stability in a time efficient manner.

\section{Superquadric Representation} \label{sec:superquadric representation}

A SQ can be defined as a generalized quadric in which the exponents of the implicit representation of the surface are arbitrary real numbers, allowing for a more flexible set of shapes while keeping the symmetric characteristics of the regular quadric.

The superellipsoid belongs to the family of SQ; other members of the family are the superhyperboloid and the supertoroid. A SQ can be obtained as the spherical product of two superellipses \cite{1673799}, $S_1$ and $S_2$, to obtain the parametric equation

\begin{equation}
r(\eta, \omega) = S_1 \big(\eta) \oplus S_2 \big(\omega) =\left(\begin{array}{c} a_1\cos^{\epsilon_1} \eta \cos^ {\epsilon_2} \omega \\ a_2\cos^{\epsilon_1} \eta \sin^ {\epsilon_2} \omega \\a_3\sin ^{\epsilon_1} \eta \end{array}\right),
\label{eq:superParam}
\end{equation}
with $\eta  \in[\frac{-\pi}{2},\frac{\pi}{2}]$ and $\omega \in[-\pi, \pi]$,
where $a_1$, $a_2$, $a_3$ are the scaling factors of the three principal axes. The exponent $\epsilon_1$ controls the shape of the SQ's cross-section in the planes orthogonal to $(x,y)$ plane, and $\epsilon_2$ controls the shape of the SQ's cross-section parallel to the $(x,y)$ plane. The pose of the SQ with respect to a world frame is specified by the six parameters that define a rigid motion, $p_x$, $p_y$, $p_z$ for the position vector and $\rho$, $\psi$, $\theta$ for defining a rotation matrix, for instance using roll, pitch, and yaw angles. The total set of parameters that fully defines the SQ consists of 11 variables, \{$a_1$, $a_2$, $a_3$, $\epsilon_1$, $\epsilon_2$, $p_x$, $p_y$, $p_z$, $\rho$, $\psi$, $\theta$ \}.  

The SQ can also be expressed using an implicit equation in normal form as
\begin{equation}
f(a,x,y,z): \quad \bigg(\bigl\lvert\frac{x}{a_1} \bigl\lvert^\frac{2}{\epsilon_2} +\bigl\lvert\frac{y}{a_2} \bigl\lvert^\frac{2}{\epsilon_2} \bigg)^\frac{\epsilon_2}{\epsilon_1} + \bigl\lvert\frac{z}{a_3}\bigl\lvert^\frac{2}{\epsilon_1} = 1.
\end{equation}
It is easy to show that the SQ is bounded by the planes given by $-a_1 \le x \le a_1$, $-a_2 \le y \le a_2$, and $-a_3 \le z \le a_3$.

In \cite{Pilu:1995:ESS:236190.236215}, a method is proposed for uniform spatial sampling of points on SQ models. The arc length between two close points on $x(\theta)$ and $x(\theta + {\Delta_\theta}\theta)$ on a curve can be estimated by Euclidean distance,
\begin{equation}
D(\theta) = |x(\theta)-x(\theta + {\Delta_\theta}\theta))|
\end{equation}
where we can use a first-order approximation for ${\Delta_\theta}\theta$,

\begin{equation}
{\Delta_\theta}\theta = \frac{D(\theta)}{\epsilon}* \sqrt{\frac{\cos(\theta)^2 \sin(\theta)^2}{a_1^2\cos(\theta)^{2\epsilon}\sin(\theta)^4+a_2^2\sin(\theta)^{2\epsilon}\cos(\theta)^4}}
\end{equation}

By setting $D(\theta)$ to a constant sampling rate, the incremental updates of the angular parameters are

\begin{equation}
\theta_i = \begin{cases}
 \theta_{i-1} + \Delta_\theta(\theta_i), & \theta_0 = 0, i\in\{1....N\}, \theta_N < \frac{\pi}{2}, \\
  \theta_{i-1} - \Delta_\theta(\theta_i), & \theta_0 = \frac{\pi}{2}, i\in\{1....N\}, \theta_N >0,
\end{cases}
\end{equation}

where
\begin{equation}
{\Delta_\theta}\theta = \bigg(\frac{D(\theta)}{a_2}-\theta^\epsilon\bigg)^\frac{1}{\epsilon} - \theta
\end{equation}
for $\theta\rightarrow0$, and
\begin{equation}
{\Delta_\theta}\theta = \bigg(\frac{D(\theta)}{a_1}-(\frac{\pi}{2}-\theta)\bigg)^\frac{1}{\epsilon} - (\frac{\pi}{2}-\theta)
\end{equation}
for $\theta\rightarrow\frac{\pi}{2}$.

This approach has been implemented for the SQ sampling in our work.

\section{Partial-view Model Completion}\label{sec:model completion}
\begin{figure}[h!]
\centering
\includegraphics[scale=0.35]{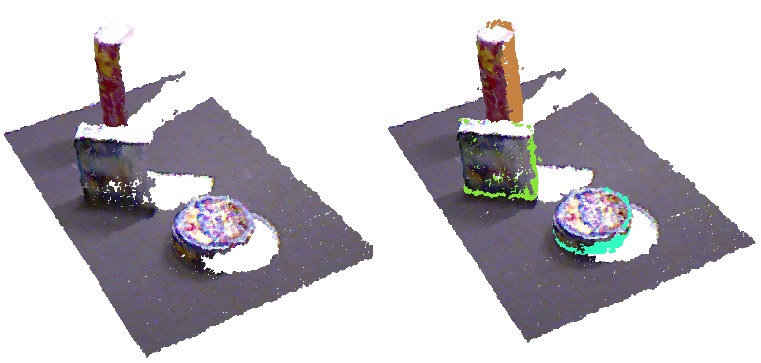}
\caption{Point cloud mirroring on individual objects of the scene. From left to right a) point cloud of three objects on the table (the depth cloud capturing device is at the left of the image), b) the point cloud is mirrored and the approximated complete model is constructed. The mirrored points for different objects are shown in distinct colors.}
\label{fig:mirrored_scene}
\end{figure}

In this section, the procedure for completion of object models from the partial view point cloud is described. A typical point cloud view from a point cloud capturing device is shown in Fig.  \ref{fig:mirrored_scene}a, where only a part of the object is captured. The other parts of the object are on the occluded side, so they can only be captured by a device situated on the other side of the table. Reconstruction of the full object model can then be performed by registering the two views from two individual devices. As most of the objects in our daily lives are symmetric by nature, by the law of symmetry it can be asserted that the occluded side of the object is merely a reflection of the visible part. By using only this assumption, the mirroring algorithm presented here creates the occluded part of the point cloud and reconstructs the object model online.

The input to the system is an organized point cloud captured from a single view 3D point cloud capturing camera. The scenes can be captured by an RGB-D camera such as the Kinect, time-of-flight cameras, or stereo cameras. In our experiments, a single Kinect camera is used. The process is divided into three steps, a first segmentation step, second pose estimation step, and finally a mirroring step.

\subsection{Segmentation}
\begin{figure}[h!]
\centering
\includegraphics[scale=0.24]{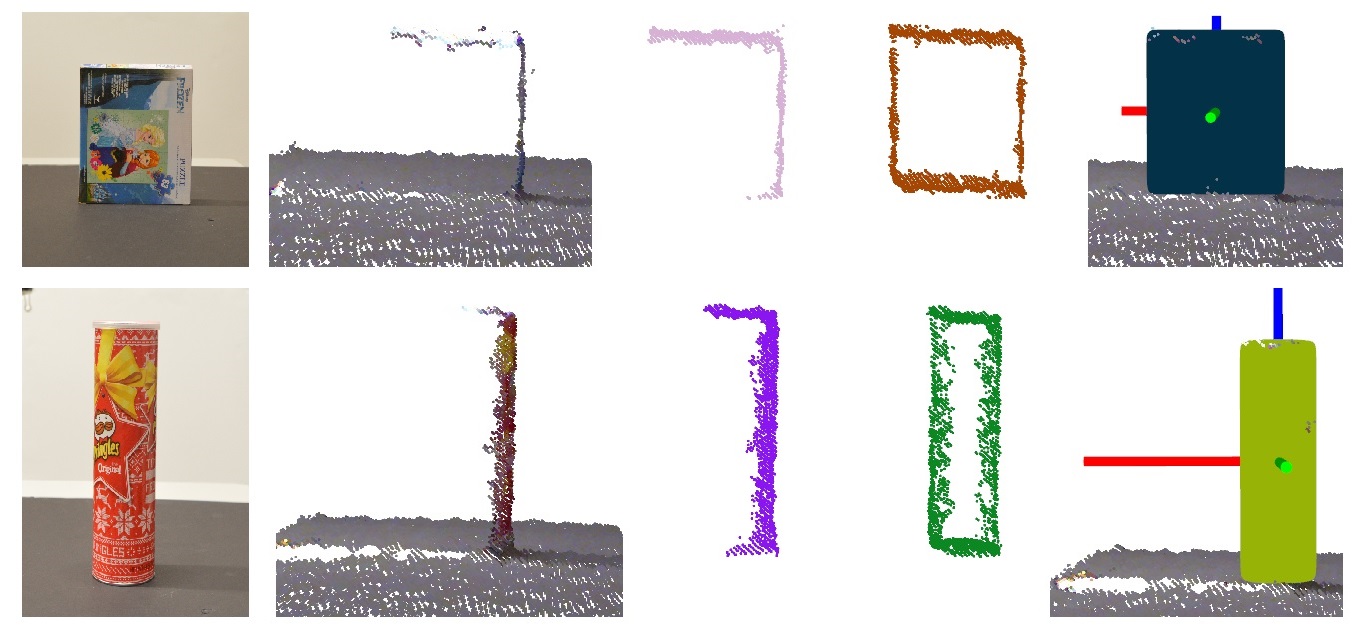}
\caption{The complete mirroring pipeline. From top to bottom, the mirroring process with a box and a cylinder object (The capturing device can be assumed at the right side of the image): a) RGB side view of the object, b) captured point cloud of the scene, c) segmented point cloud of the object, d) mirrored point cloud with the segmented cloud, e) SQ representation of the object.}
\label{fig:mirroring}
\end{figure}

The initial scene contains objects situated on a table plane. After the dominant table plane is removed, we assume that the clusters present on the table plane correspond to objects. The Voxel Cloud Connectivity Segmentation (VCCS) \cite{6619108} algorithm is used to separate the clusters. The 3D point clouds can be over-segmented into surface patches, which are considered as supervoxels. The algorithm utilizes a local-region growing variant of K-Means Clustering to generate individual supervoxels, represented as $\bar p_i = (\bar x_i, \bar n_i, N_i)$, where $\bar x_i$ is the centroid, $\bar n_i$ is the normal vector and $e \in N_i$ are the edges to adjacent supervoxels. The connections of supervoxels are checked for convexity and those belonging to same connectivity graph are segmented into individual clusters. The results of the segmented point clouds are visualized in Fig. \ref{fig:mirroring}c.

\subsection{Pose estimation}
Estimating correct pose is one of the most important characteristics of any grasping scenario. The most common approach to estimate pose from point clouds where ground truth orientation of the object model is not known is by using Principal Component Analysis (PCA) for orientation and centroid information for positions. The dominating eigen vectors from PCA are considered as the orientation of point clouds. But this approach utilizes the point cloud on the surface of the object.  While the centroid from these methods may be the center of the point cloud, it is not the centroid of the object. Also, to deal with noisy point clouds from real sensors, the axes cannot be determined from PCA, as for every iteration, the number of points perceived varies rapidly. 

\begin{algorithm}  
\caption{Generate Mirrored Cloud} input (organized point cloud $Cloud$)
\label{mirror_algo}
\begin{algorithmic}[1]
\Procedure{$CompleteCloud$} {}
\\Segment out the Table
\\Segment Objects
\For{each $i$ in $Objects$   }
\State Find pose ($c_x, c_y, c_z$, $\rho$, $\psi$, $\theta$)  \
\State Generate local frame at $c_x, c_y, c_z$
\For {each point $p$ in $i$}
\State Calculate Euclidean distance ($c,p$)
\State Negate distance and find point $m$
\State Add $m$ to $i$
\EndFor
\EndFor
\EndProcedure
\end{algorithmic}
\end{algorithm}

The center of the object ($c_x, c_y, c_z$) can be estimated by taking the mean of the values of the points in all three dimensions. For completely unknown objects, the ground truth orientation is also unknown, so an assumption is asserted that the orientation of all the objects perceived are $z$-axis upwards. The problem of obtaining pose information ($\rho$, $\psi$, $\theta$) is now limited to finding only the $\theta$ which control the rotation in $z$-axis. The volume $V_{cloud}$ can be calculated by maximum distances in each dimension($d_x$, $d_y$, $d_z$), where the maximum distance is the difference between the maximum value and the minimum value in each plane. All the points in the point cloud are projected onto the table plane and a 2D convex hull is created around the projected cloud. The volume of a bounding box can also be estimated from length $l$ and width $w$ of the convex hull and $h$ as the value of $p_{Zmax}$. By rotating the convex hull in small iterations and minimizing error ($\epsilon$) between the volume of the bounding box and the volume of the point cloud,  the rotation $\theta$ in $z$-axis can be obtained.

\begin{equation} \label{min_volume}
\epsilon = min(V_{cloud} - [l*d*h])
\end{equation}

\subsection{Creating a mirrored surface}
To obtain a complete approximated model of the perceived object, the points perceived are mirrored based on the pose ($c_x, c_y, c_z$, 0, 0, $\theta$) found in the previous step. A local reference frame is assumed at the center of the object and the euclidean distance of all the points from the centroid is calculated. All the points are reflected on the occluded side by scaling all of them by the negative value of 1 unit to the distance. So for every single point $P_i$ ($p_x, p_y, p_z$), a mirrored point $M_i$ ($-p_x, -p_y, -p_z$) is created where $p_x$, $p_y$, $p_z$ are distances in $x, y, z$ directions from the center of the object. Though the pose does not contain any $\rho$ and $\psi$ values, this mirroring method can work on any arbitrary orientation of the object. To deal with noisy data, all the point clouds are processed through a series of outlier removal algorithms.

As shown in Fig. \ref{fig:mirrored_scene}b, the objects are captured from a point of view (POV) where only a part of the object can be seen. After removing the points belonging to the table surface and segmenting out the object of interest from the environment, the object point clouds are depicted in Fig. \ref{fig:mirrored_scene}c. The mirrored point cloud with object points are shown in Fig. \ref{fig:mirrored_scene}d. It is worth noticeable that the pose of the box in Fig. \ref{fig:mirrored_scene} is such that only the smallest part of the box can be perceived, but the algorithm still creates the approximate model of the box. A scene with mirrored point clouds are shown in Fig. \ref{fig:mirrored_scene}, where the points with distinct colors are mirrored from the object point cloud by the mirroring algorithm presented in Algorithm \ref{mirror_algo}.

\section{Superquadric fitting}\label{sec:fitting}
To fit the SQ model to point cloud data, distance from the points to the function $f(a, x, y, z)$ must be minimized. Radial distance $|OP|$, which is the distance between the points and the center of the SQ \cite{Jaklic:2000:SRS:358353}, is used instead of Euclidean distance,

\begin{equation}
d(a,x,y,z) = \min\sqrt{a_1 a_2 a_3\sum_{k=0}^{n}||OP||*(f(a,x,y,z)^\epsilon-1)}
\end{equation}

\section{Superquadric Grasp Synthesis}\label{sec:grasp_synth}
This work focuses on the use of two-fingered grippers for grasping the objects. It is well known that antipodal grasps can be successful in both convex and concave objects in the presence of friction. Finding antipodal points on generally-shaped object models usually returns many candidates, which are ranked according to some metric. In order to make the grasp more stable and balanced, we impose two conditions: a) grasping closer to the centroid of the object and b) at the points of minimum curvature. In addition, we have constraints given by the depth and width of the robotic gripper and the task-based direction of approach.

\subsection{Minimum curvature contact points}
The gripper contact points for antipodal grasping with minimum curvature can be calculated from the SQ parameters to be
\begin{equation}
p_W = [T_{WK}][T_{KS_i}](\pm \begin{Bmatrix}p_a \\ 1 \end{Bmatrix}),
\end{equation}
where $[T_{WK}]$ is the $4\times 4$ homogeneous transformation from the world frame to the camera frame, $[T_{KS_i}$ is the $4\times 4$ homogeneous transformation from the camera frame to the principal frame of SQ $i$, and $p_a$ is the position vector of the contact points in the SQ frame, to be selected according to the criteria explained below and summarized in Table \ref{tab:vectorCP}.

\begin{figure}[h!]
\centering
 \includegraphics[scale=0.03]{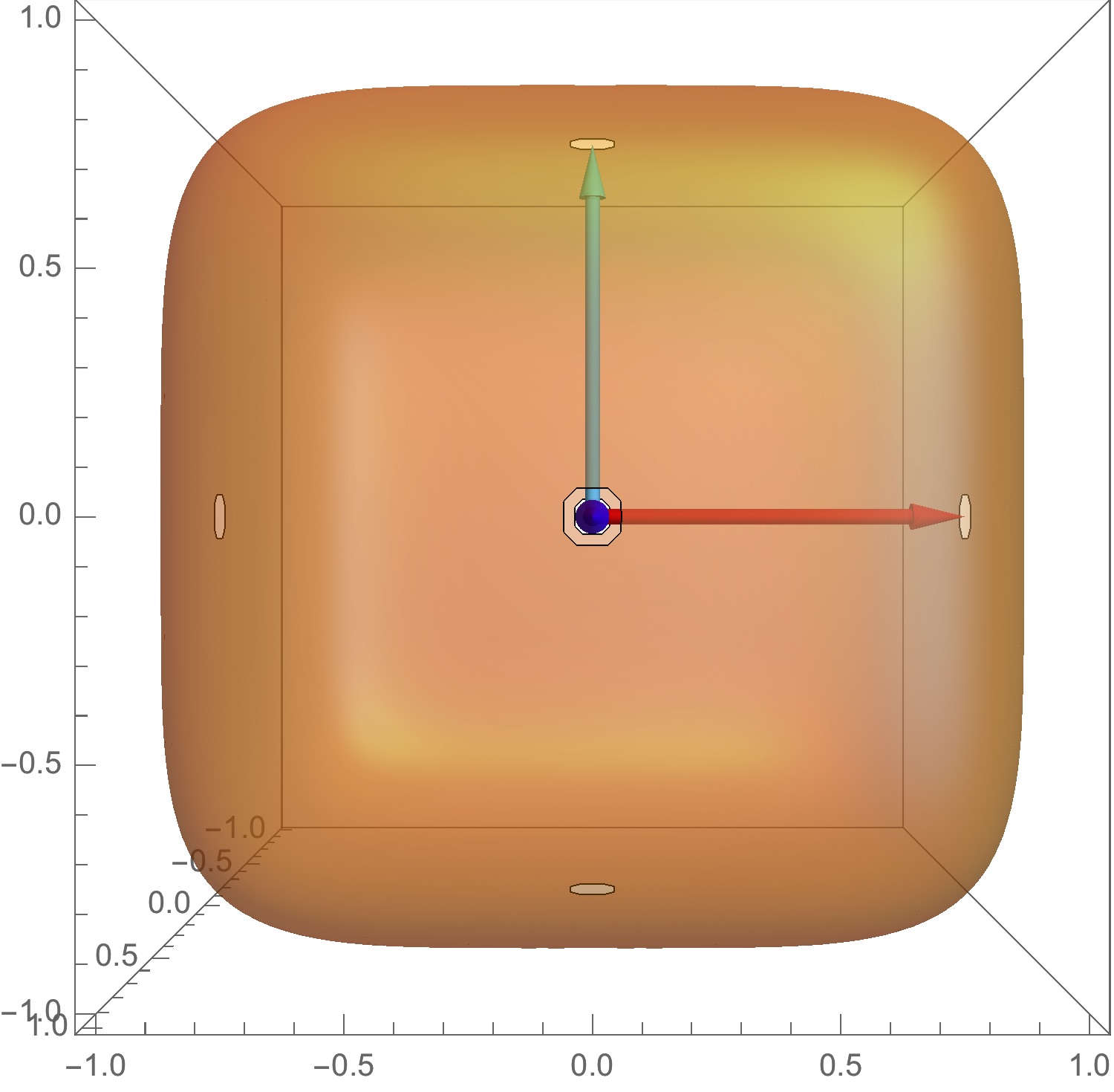}
 \includegraphics[scale=0.03]{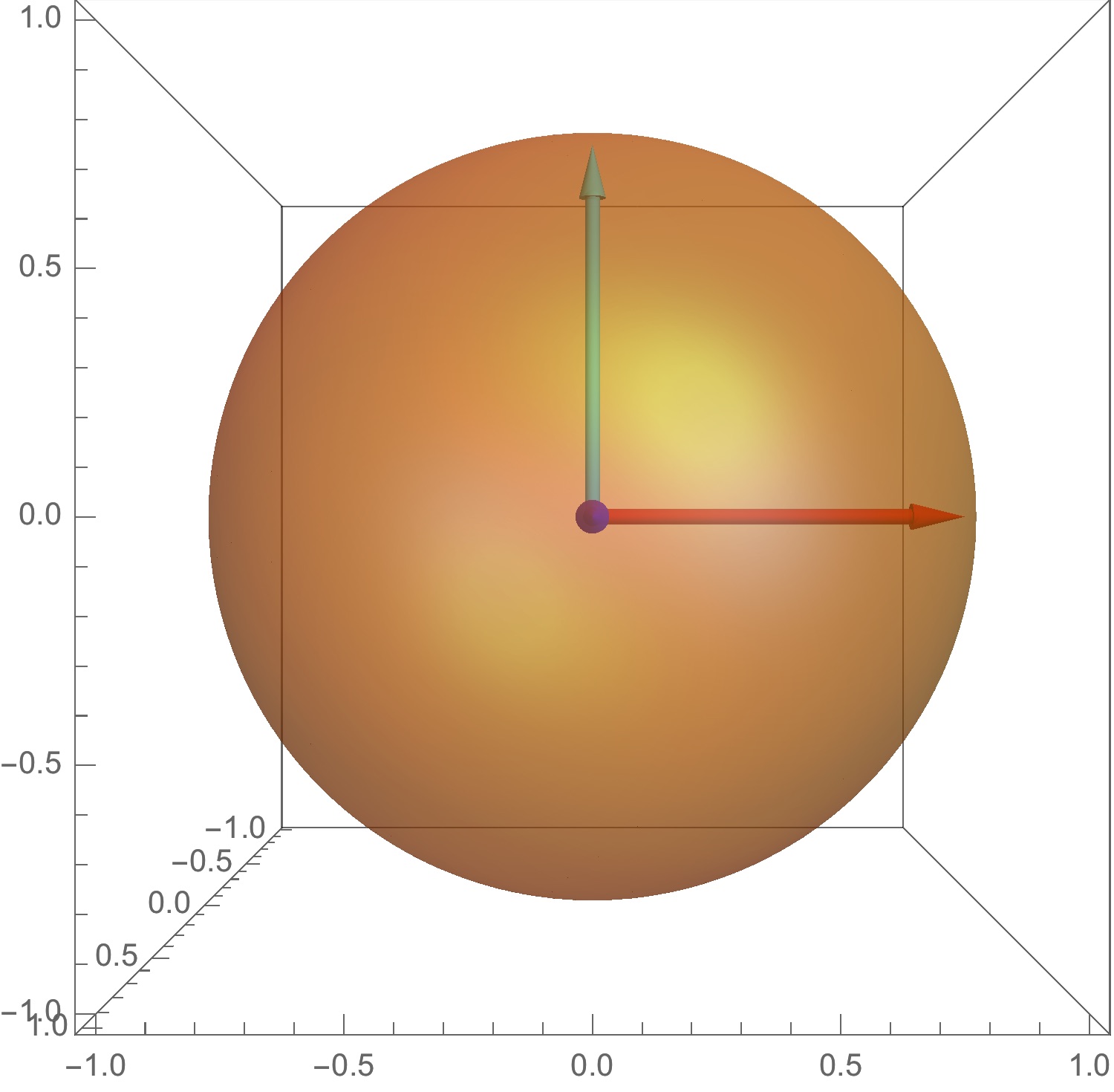}
 \includegraphics[scale=0.13]{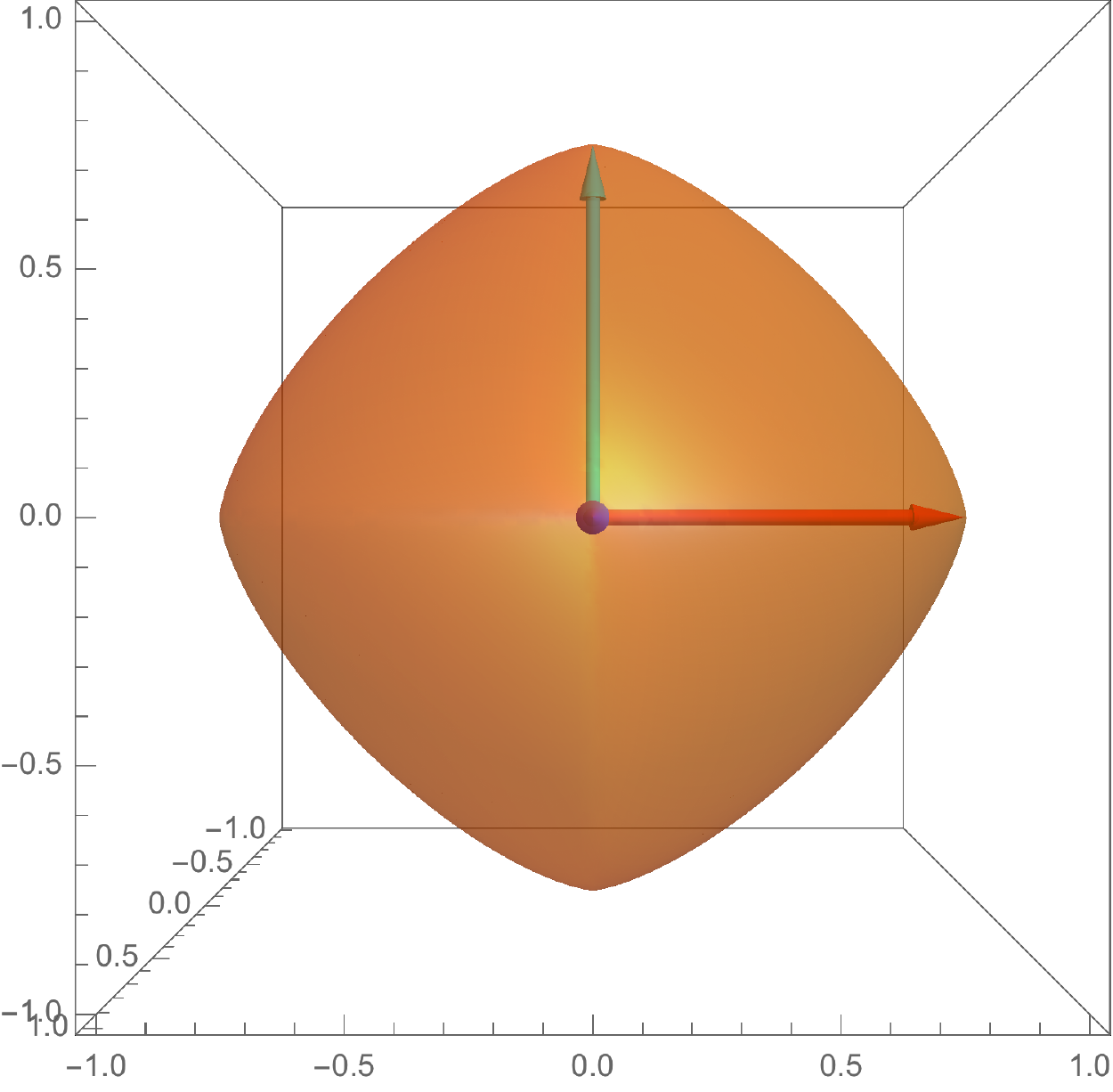}
 \includegraphics[scale=0.13]{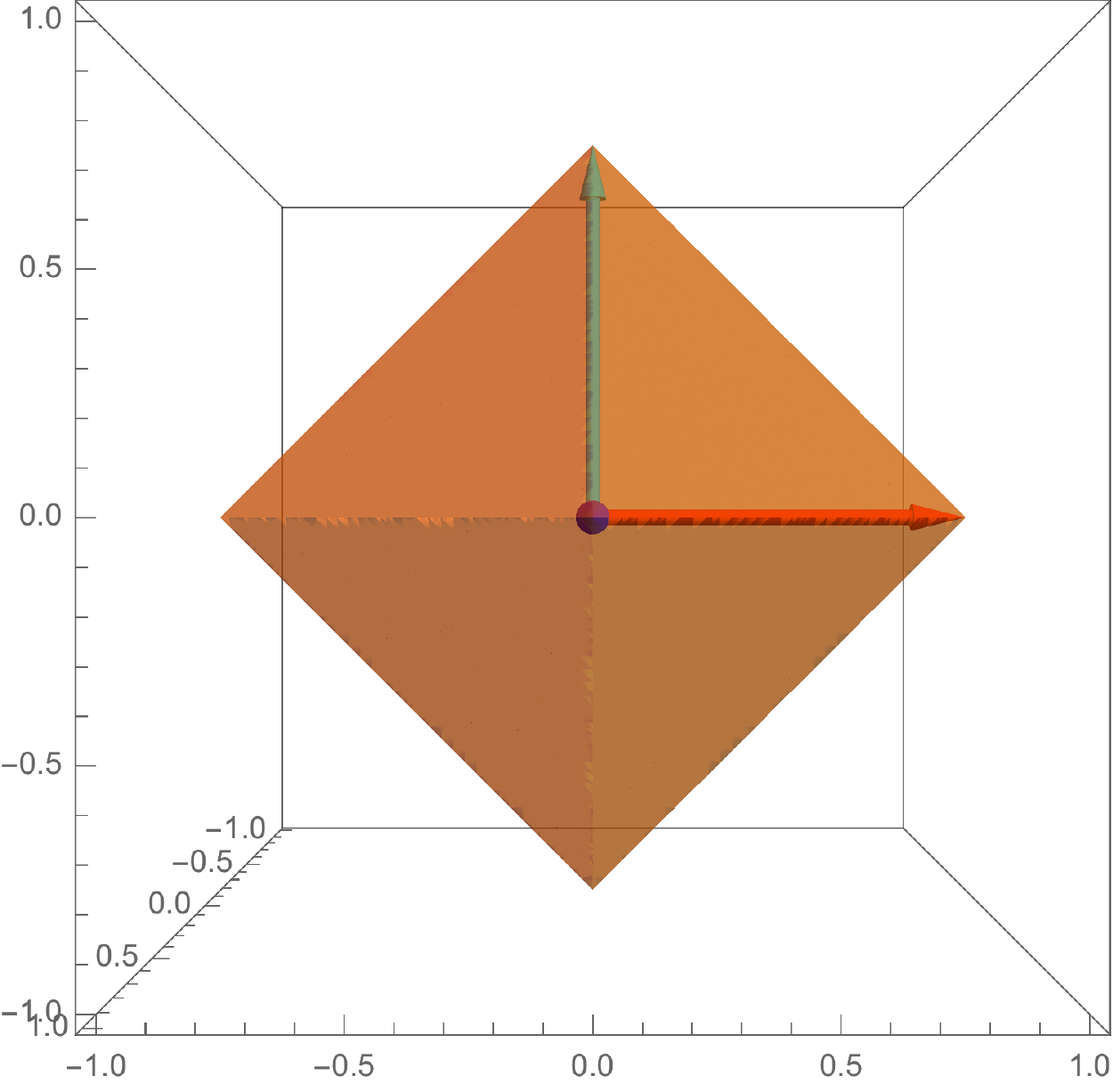}
 \includegraphics[scale=0.13]{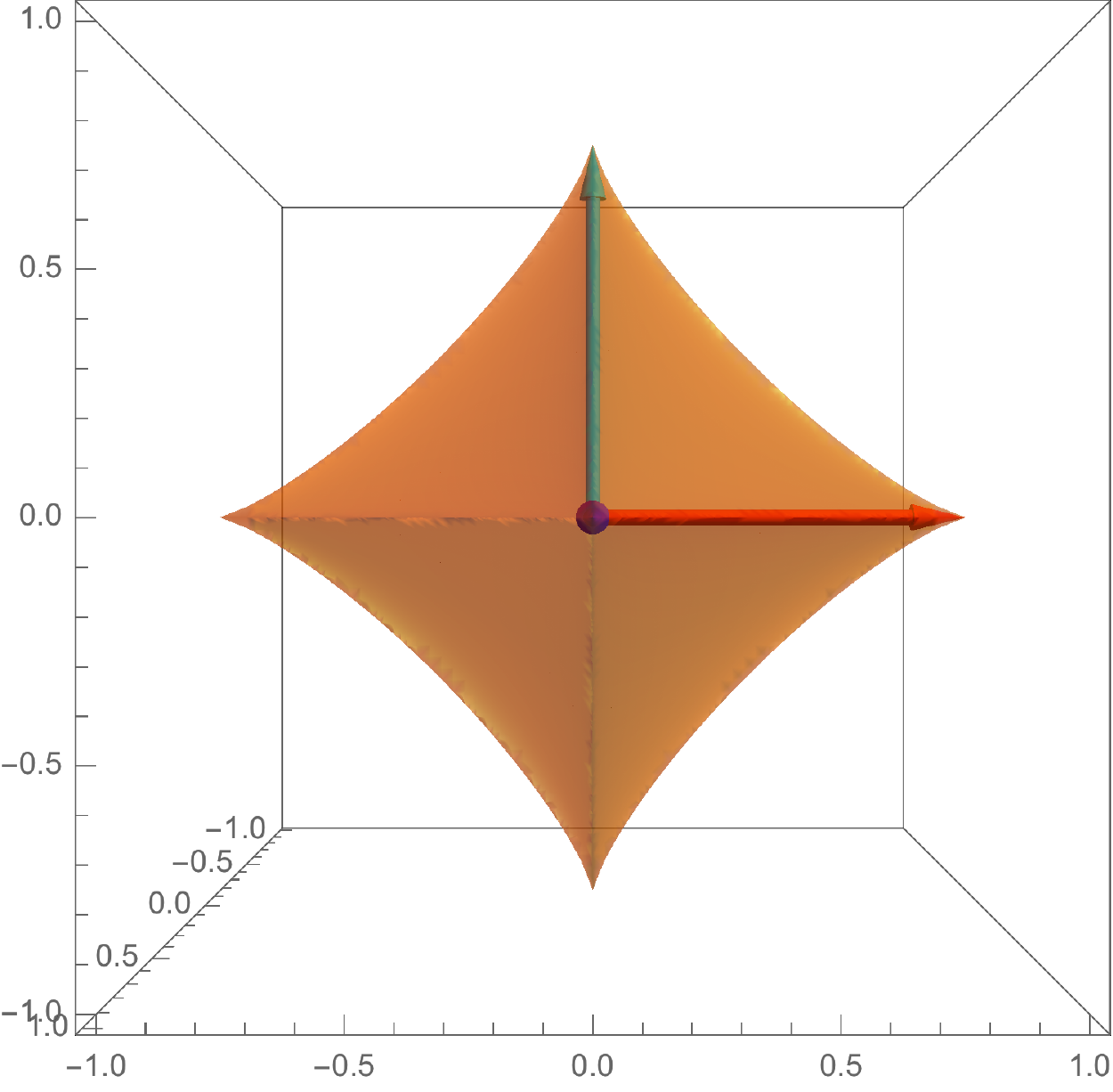}
 \caption{Convex SQ for $\epsilon=0.4$, $\epsilon=1$, $\epsilon=1.4$ and flat for $\epsilon=2$; concave SQ for $\epsilon = 2.4$.}
 \label{fig:concavity}
 \end{figure}

The SQ is convex for values of the exponent $0 < \epsilon < 2$; for larger values, it becomes concave at the corresponding cross section, as shown in Fig.\ref{fig:concavity}.

\begin{figure}[h!]
\centering
\includegraphics[scale=0.24]{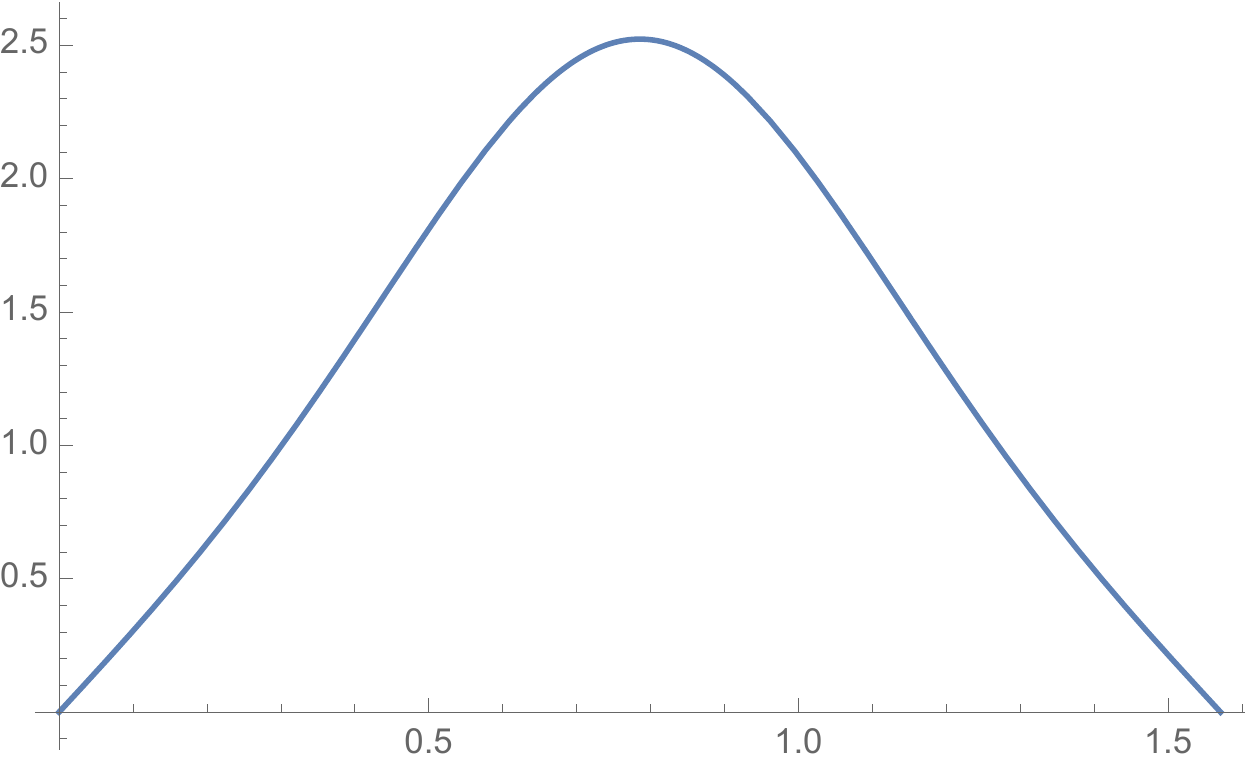}
\includegraphics[scale=0.24]{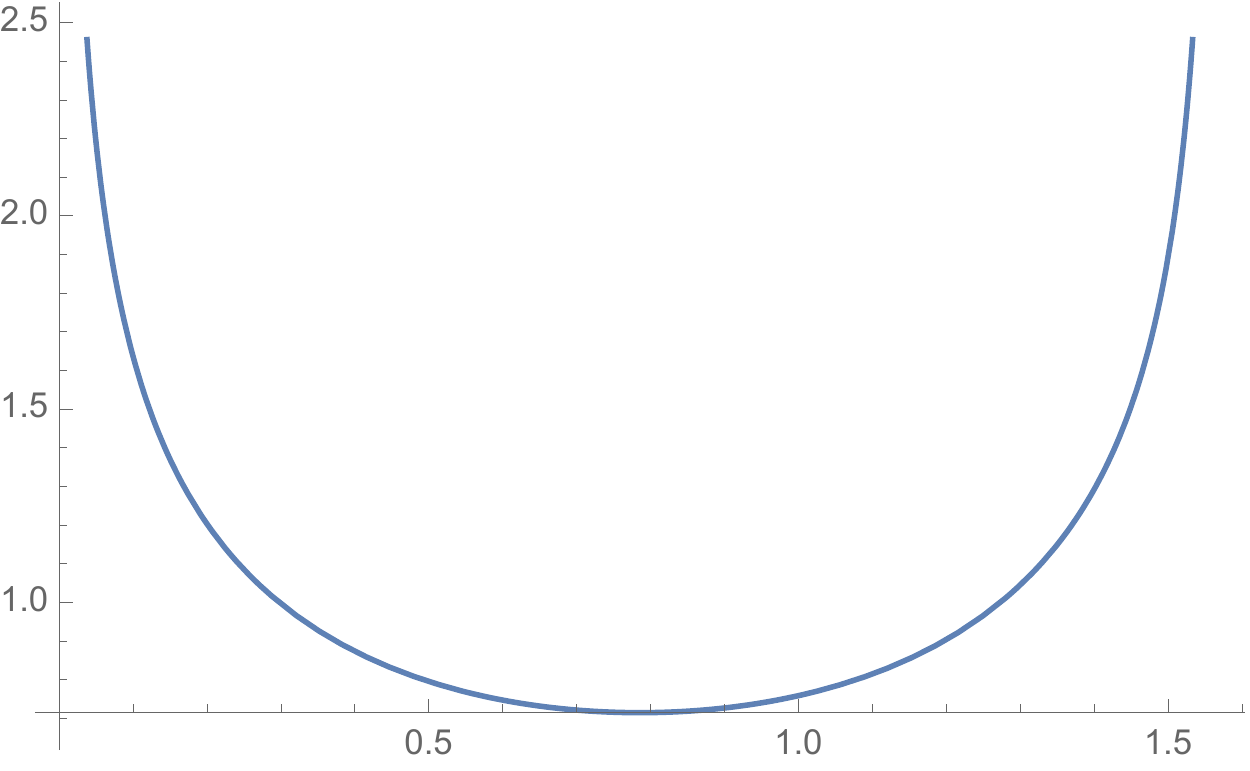}
\includegraphics[scale=0.05]{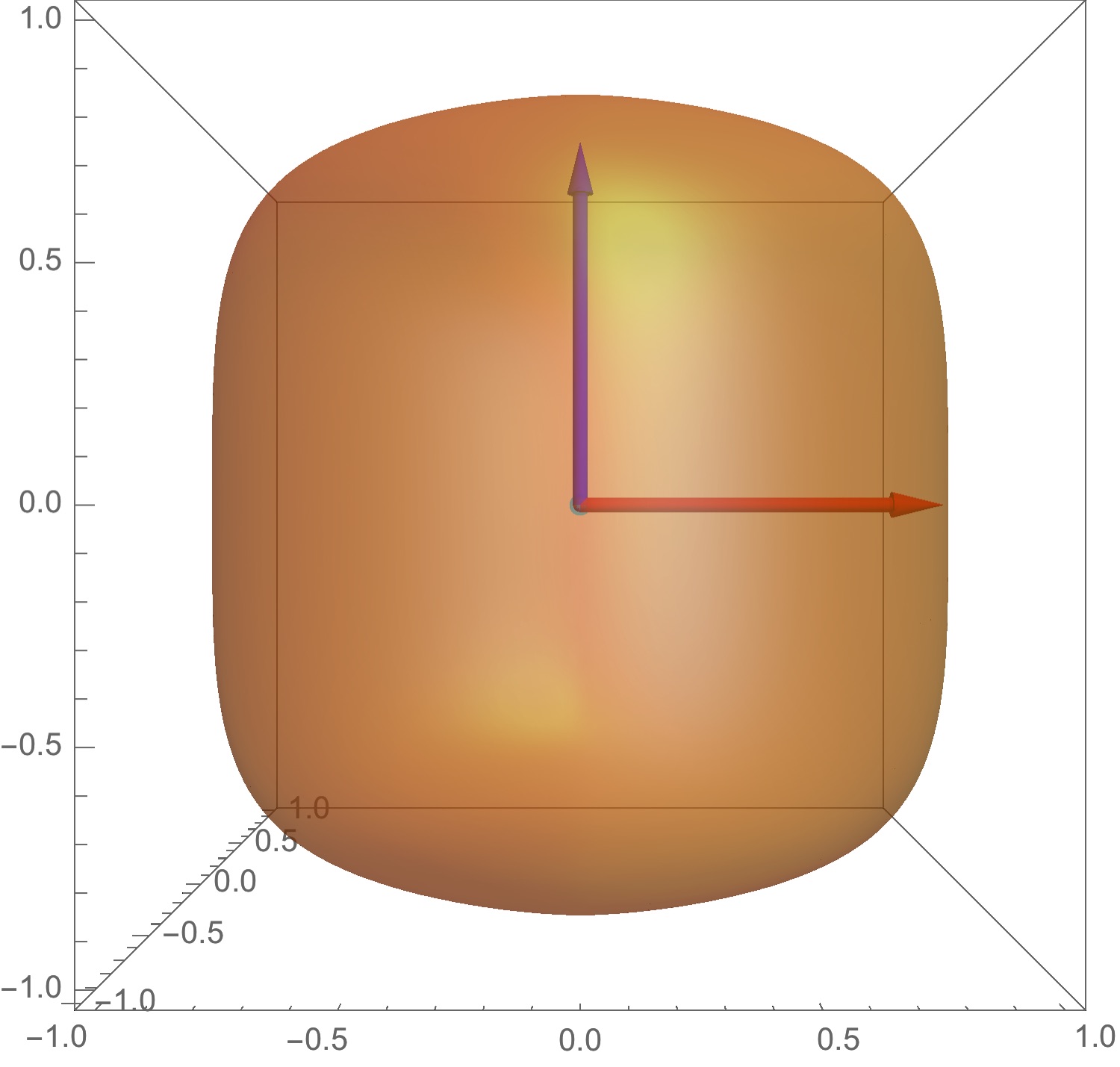}
\includegraphics[scale=0.05]{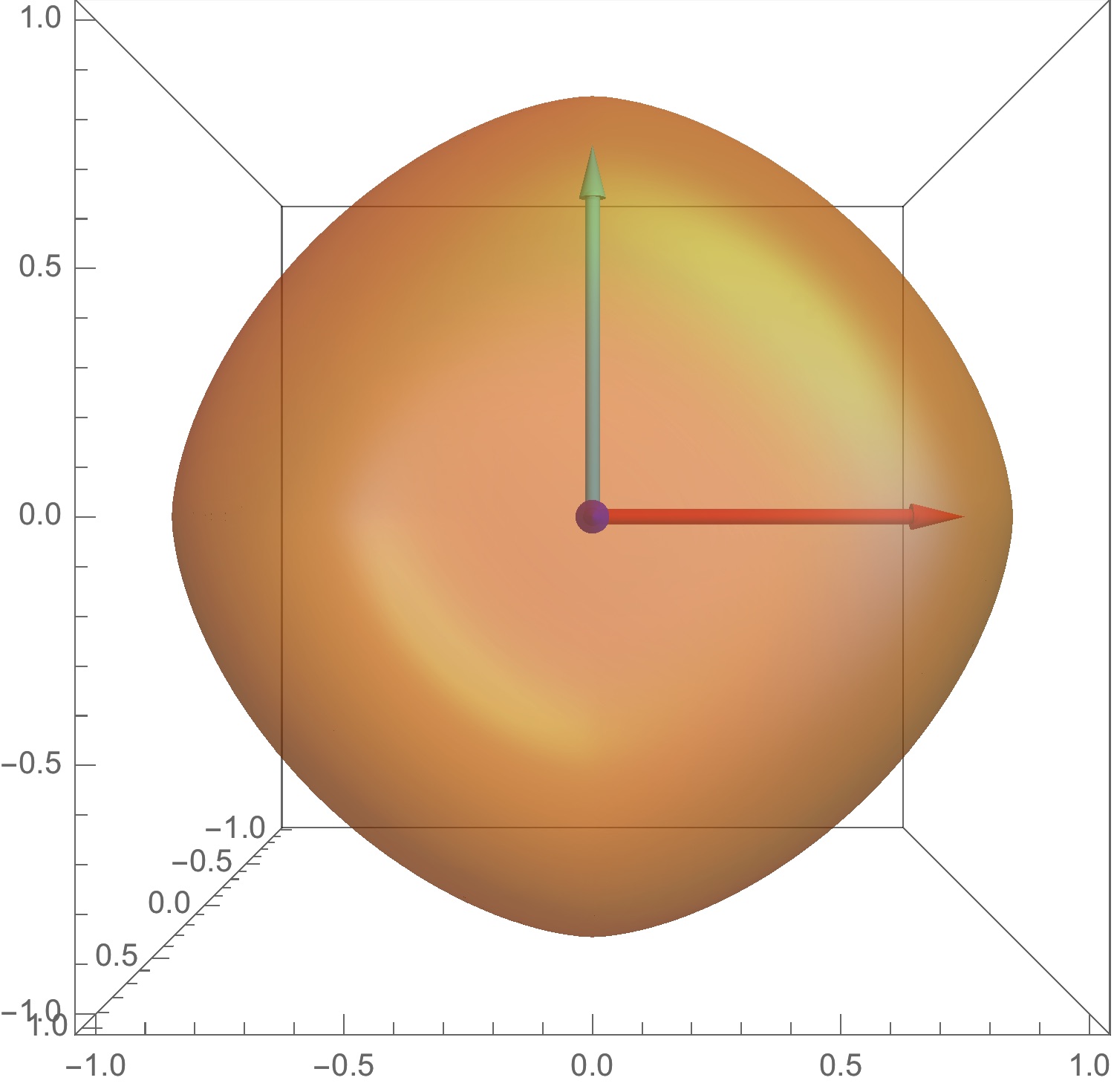}
\caption{From top to bottm: a) curvature as a function of the angular parameter and the value of the exponent $\epsilon$ in curvature plot, b) cross-sectional cut of the SQ for same semi-axes length and values $\epsilon=0.5$ and $\epsilon=1.2$.}
\label{fig:curvatures}
\end{figure}

Curvature in SQ can be easily calculated from either their implicit or their parametric expression. Considering the implicit expression in eq (\ref{eq:superParam}) at the $x-z$ plane obtained when $\omega = 0$, the curvature is
\begin{equation}
k(\eta) = \frac{(\epsilon_1-2)a_1a_3\cos^{\epsilon_1-4}\eta\sin^{\epsilon_1-4}\eta}{\epsilon_1\sqrt{(a_1^2\cos^{2\epsilon_1-4}\eta+a_3^2\sin^{2\epsilon_1-4}\eta)^3}},
\end{equation}
and similarly for the other two planes passing through the origin of the SQ's frame. Depending on the value of the exponent $\epsilon_1$, the minimum curvature will be found at the intersection with the axes or at $45^o$ from them. Fig.\ref{fig:curvatures} shows the effect of the exponent on the location of the minimum curvature; the curvature is constant for $\epsilon_1=1$.

For substantially different lengths of the semi-axes of the convex SQ, the points of minimum curvature at $\epsilon > 1$ are not antipodal. Here, normal directions to the SQ are found with minimum angular deviation. Disregarding its sign, the normal to the SQ can be calculated for a coordinate plane passing through the origin as the second derivative of the parametrized curve. For $x-z$ plane at $\omega=0$ normal $n$ is

\begin{equation}
n(\eta) = \begin{Bmatrix}
\frac{a_3 \cos^{\epsilon_1+2}\eta \sin^{2\epsilon_1}\eta}{\sqrt{a_3^2\cos^{2\epsilon_1+4}\eta\sin^{4 \epsilon_1}\eta+a_1^2\cos^{4\epsilon_1}\eta\sin^{2 \epsilon_1+4}\eta}} \\ 
\frac{a_1 \cos^{2\epsilon_1}\eta \sin^{\epsilon_1+2}\eta}{\sqrt{a_3^2\cos^{2\epsilon_1+4}\eta\sin^{4 \epsilon_1}\eta+a_1^2\cos^{4\epsilon_1}\eta\sin^{2 \epsilon_1+4}\eta}} 
\end{Bmatrix}
\end{equation}

\begin{figure}[h!]
\centering
\includegraphics[angle=90,scale=0.45]{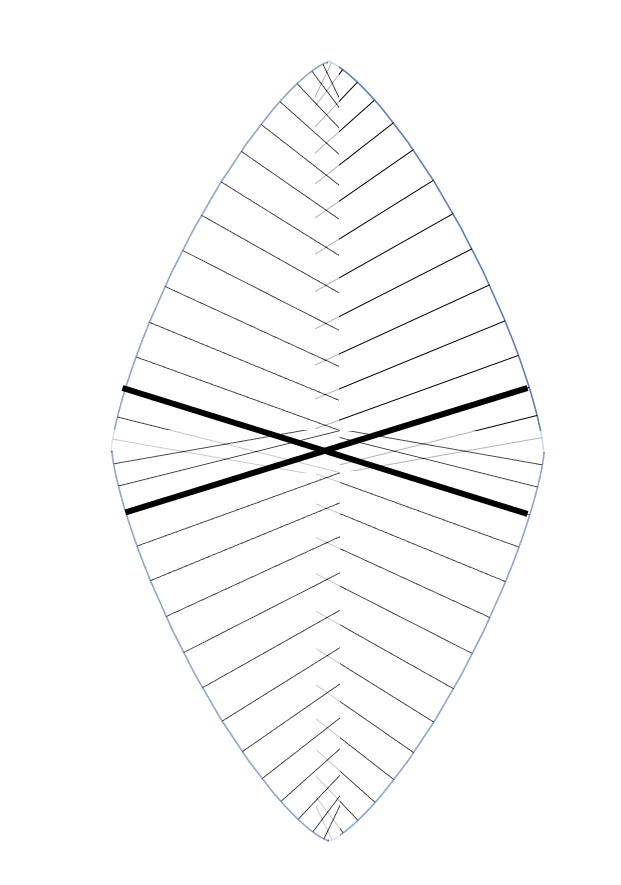}
\caption{Normals to the surface of the SQ for $a_1=0.5$, $a_3=0.9$ and $\epsilon_1 = 1.5$. Normal lines of approximate antipodal points marked with thicker lines}
\label{fig:normalAnti}
\end{figure}

Fig. \ref{fig:normalAnti} shows an example of the antipodal points with minimum curvature on the $x-z$ plane; if a solution exists, then there will always be another symmetric solution. In addition, the contact points at the major axes are antipodal, but depending on the value of the exponent, they may present higher curvature or a cusp for extreme cases.

The angular parameter of the approximate antipodal points in the plane can be found by imposing that the normal line passes close enough to the origin,
\begin{equation}
p_a = \min_{\eta}(\frac{a_1a_3 \cos^{2\epsilon_1}\eta \sin^{2\epsilon_1}\eta}{\sqrt{a_3^2\cos^{2\epsilon_1+4}\eta\sin^{4 \epsilon_1}\eta+a_1^2\cos^{4\epsilon_1}\eta\sin^{2 \epsilon_1+4}\eta}} -1).
\label{eq:antiNormal}
\end{equation}
 
For concave SQs, the curvature helps make the grasp closed. Antipodal points are sought using eq (\ref{eq:antiNormal}).

\begin{table}[h!]
\caption{Position vector to contact points as a function of the length of the semi-axis and the exponent $\epsilon$.}
\begin{center}
\begin{footnotesize}
\begin{tabular}{ |c|c|c| }
\hline
Exponent & Dimensions & Position vector\\ 
\hline
$\epsilon < 1$ & $a_1 > a_2$ & $p_a=(0,a_2,0,1)$ \\ 
 & $a_1 = a_2$ & $p_a=(a_1, 0,0,1)$ or  $p_a=(0,a_2,0,1)$\\ 
  & $a_1 < a_2$ & $p_a=(a_1,0,0,1)$  \\ 
  \hline
$\epsilon = 1$ & $a_1 > a_2$ & $p_a=(0,a_2,0,1)$ \\ 
 & $a_1 = a_2$ & $p_a=(a_1\cos\omega,a_1\sin\omega,0,1)$ for all $\omega$  \\ 
   & $a_1 < a_2$ & $p_a=(a_1,0,0,1)$  \\ 
 \hline
 $\epsilon >1$  & $a_1 = a_2$ & $p_a=(a_1/\sqrt{2},a_1/\sqrt{2},0,1)$ \\ 
  & $a_1 \ne a_2$ & $p_a(\omega)$  as per Eq.(\ref{eq:antiNormal}) \\
\hline
  \end{tabular}
  \label{tab:vectorCP}
  \end{footnotesize}
  \end{center}
\end{table}

\subsection{Direction of approach}
Let us define the plane of grasping as the plane containing the contact points and normal to the fingers of the gripper. The direction of approach of the gripper is then perpendicular to this plane.  Because of the condition of entered grasp, at least one of the dimensions of the superellipsoid at the plane of grasping, $a_i$, needs to be smaller than the width of the gripper, $w$. In addition, the dimension of the major axis in the direction perpendicular to that plane needs to be smaller than the depth of the gripper, $h$. 

\begin{figure}[h!]
\centering
\includegraphics[scale=0.35]{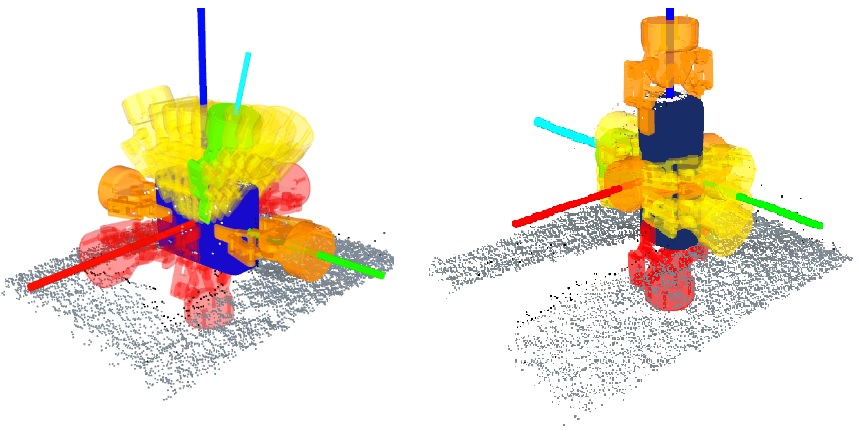}
\caption{Grasps on the (SQ). From left to right, a) on a box, b) on a cylinder. The red grasps are invalid, yellow grasps are potential grasp on different axes, light yellow grasps are potential grasp on current axes and green grasp in the best grasp. The approach direction is shown by cyan bar.}
\label{fig:show_grasps}
\end{figure}

\begin{figure*}
\centering
\includegraphics[width=\textwidth,height=6.5cm]{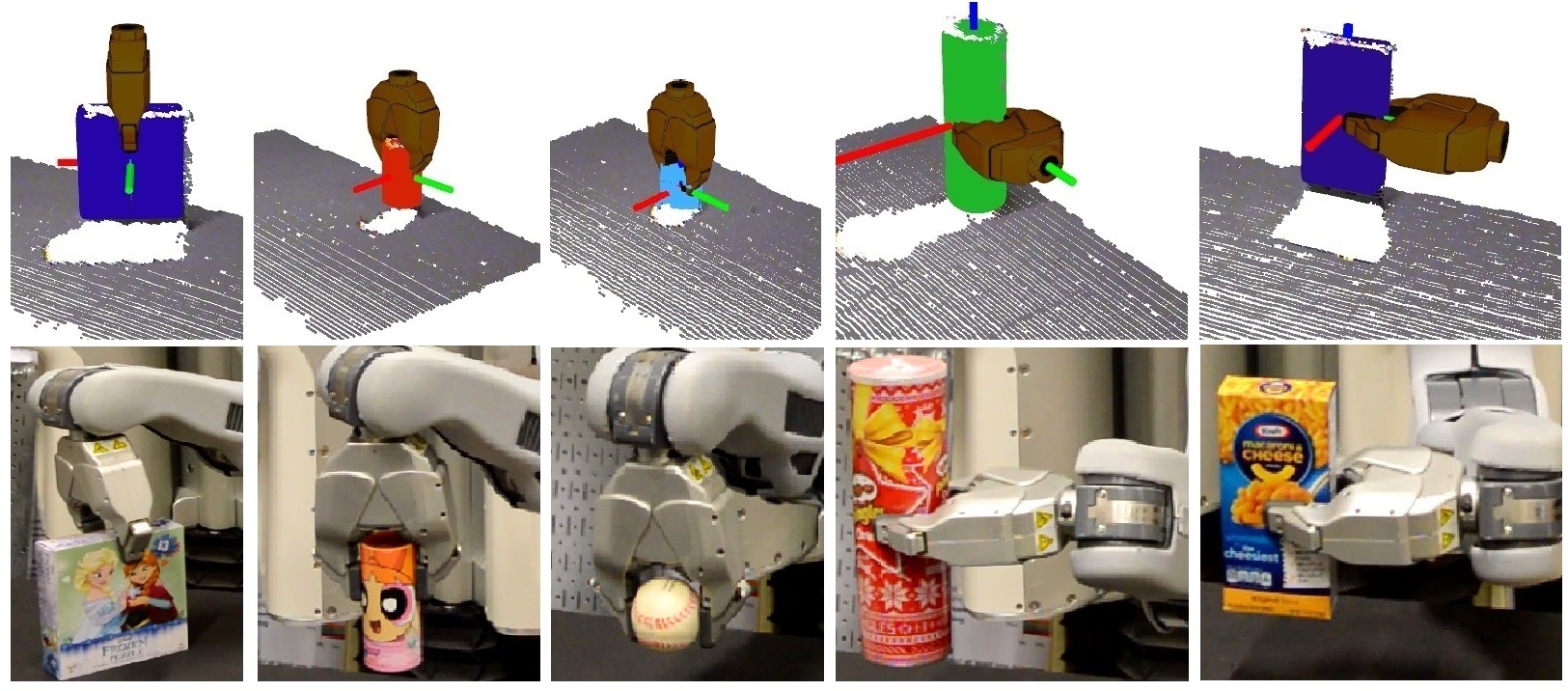}
\caption{Experiments with the PR2 robot with 5 different objects. Top: Visualization of the gripper on the obtained SQ before executing. Bottom: Grasp execution on objects in real time}
\label{fig:pr2_grasp}
\end{figure*}

The plane of grasping to consider can be at an angle from the coordinate planes in order to accommodate verticality or the above conditions. 
As before, the direction of approach is calculating as

\begin{equation}
s_W = [T_{WK}][T_{KS_i}](\pm \begin{Bmatrix}s_a \\ 0\end{Bmatrix}),
\end{equation}

where $s_a = x, y$, or $z$ or a coordinate rotation about one of those axes.

Algorithm \ref{GSAlgorithm} shows the summary of the process to select the plane of grasping and the contact points. It is intuitive that the algorithm always tries to approach object from the minimum $\pm a_i$ direction. It shown in Fig. \ref{fig:pr2_grasp}, where the first and last images shows grasping of the box from the minimum $\pm a_i$ direction.

\begin{algorithm}  
\caption{Grasp synthesis} input (SQ parameters, depth of the gripper $h$, width of the gripper $w$)
\label{GSAlgorithm}
\begin{algorithmic}[1]
\Procedure{$Select Contact Planes$} {}
\For{$s_a$ SQ direction}
Calculate $s_W$
\If {angle($s_W$,$Z_W$) $> \alpha$ or length($s_a$) $> h$}
\State{Discard $s_a$}
\EndIf
\EndFor
\For{remaining $s_a$}
\If{$2*a_i > w$ for perpendicular directions}
 \State{Discard $s_a$}
 \EndIf
 \EndFor
\EndProcedure
\Procedure{$CalculateContactPoints$}{}
\State{Select $p_a$ (Table \ref{tab:vectorCP})}
\State{Calculate $p_W$}
\EndProcedure
\end{algorithmic}
\end{algorithm}

\section{Experimental Results} \label{sec:results}

This section discusses the results of the replication and SQ fitting method, and  the results from the grasping algorithm. Our dataset consists of simple household items of various shapes: balls, cubes, paper rolls etc. For experiments, the approach direction $s_W$ in negative $z$ direction is always discarded, as by our pose estimation process, we assumed that the pose is $z$-axis upwards. So the object can not be grasped through the table. Fig. \ref{fig:show_grasps} shows the visualization of potential grasps on a box and a cylinder SQ. The red grasps are easily discarded as the object can not be grasped from those directions. The dark yellow grasps are from those directions from where the object can be grasped. By incorporating motion planning and robot reachability with grasps, we decide on the grasp to be executed. The first grasp is at the $min(a_i)$ direction. If the motion planner fails to reach the main grasp at direction $min(a_i)$, we sample the grasps with varying angles at the same direction (yellow grasps in Fig.\ref{fig:show_grasps}). If all the grasps at this direction fails, we move to second $min(a_i)$ direction. 

We have performed the grasping experiments on a PR2 robot that uses a Kinect 1 to perceive objects. A grasp is considered to be successful if the object is been picked up from the table and placed in a bin situated on the other side of the setup (refer to Fig. \ref{fig:pr2_setup} for the locations of the objects and bin). To merge the fitting process with grasping as an online system, only the five parameters \{$a_1$, $a_2$, $a_3$, $\epsilon_1$, $\epsilon_2$\} are optimized by Levenberg Marquardt\cite{citeulike:1180235} algorithm implemented by Ceras library. The other pose parameters \{$p_x$, $p_y$, $p_z$, $\rho$, $\psi$, $\theta$ \} are optimized separately by the process described above in the pose estimation section of mirroring algorithm. While the range of $\epsilon_1$ and $\epsilon_2$ is constrained between the range of $0.1$ and $1.9$, the initial values of $a_1$, $a_2$ and $a_3$ are eigenvalues of the covariance matrix of the point cloud. A cartesian planner has been implemented for the system to reach the grasp by defined approach and pick up the object. For the placing task, a joint controller is implemented. 

\begin{figure*}
\centering
\includegraphics[width=\textwidth,height=5cm]{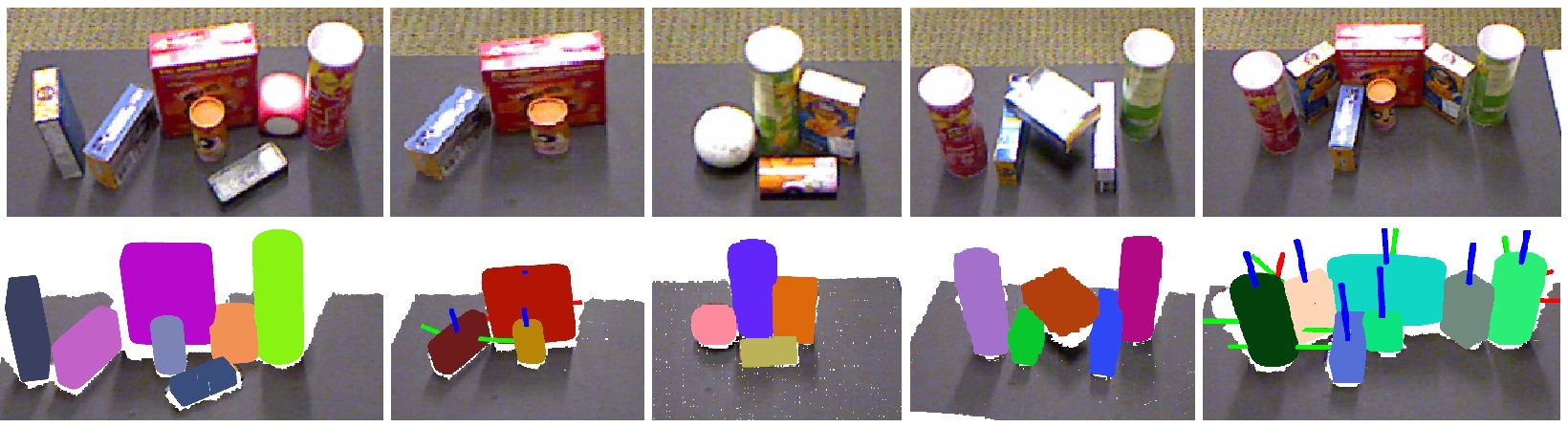}
\caption{SQ fitting in cluttered scenario. The top images show the RGB view of the image perceived by the robot. The bottom images shows the fitted SQ on those objects. The poses of the SQ are also visualized in Fig.b and Fig.e }
\label{fig:fitting}
\end{figure*}

\subsection{Superquadric fitting}
Our object dataset mostly consists of small symmetric objects, such as small boxes, cylinders and balls with varied parameters in $x,y,z$ directions. For calculating the errors, in an offline scenario, we obtained the ground truth ($a1,a2,a3$) parameters of all the objects. The $\epsilon_1$ and $\epsilon_2$ are obtained by iterating over the values between $0.1$ and $1.9$ and choosing the volume that best fits the object. The error is radial distance error between the point cloud generated by ground truth information and the point cloud generated by obtained SQ parameters. For every object, 5 different poses are defined on the table and the objects are perceived from all 5 poses. The results shown in \ref{tab:singleObject} are a comparison of recovered parameters between our method and \cite{7139713}. As the result suggests, our method shows better improved results is SQ fitting in terms of time-efficiency and accuracy, as our method determines the pose on the approximated center of the object. Also, instead of optimizing 11 parameters of the object, we only optimize 5 parameters related to object shape. While the size of the object increases, the time duration of the optimization process also increases.

\begin{table}[h]
\caption{Single object parameters. Method 1 is Mirroring and SQ Fitting, method 2 by Quispe et al. \cite{7139713}. Average time in seconds.}
\begin{center}
\begin{scriptsize}
\begin{tabular}{{|p{0.75cm}||p{0.5cm}|p{0.5cm}|p{0.5cm}|p{0.5cm}|p{0.5cm}|p{0.5cm}|p{0.5cm}|p{0.5cm}|}}
\hline
Object & Meth & $a_1$ & $a_2$   & $a_3$ & $e_1$ & $e_2$ & Avg. & Avg. \\ 
 & && &  & &  & time & error\\ 

\hline
toy & 1 & 0.023 & 0.024 & 0.15 & 0.389 & 1.031 & 0.535 & 9.2\% \\ 
cylinder & 2 & 0.022 & 0.022 & 0.15 & 0.561 & 0.928 & 1.788 & 9.78\% \\ 
\hline 
ball & 1 & 0.69 & 0.09 & 0.45 & 1.003 & 0.95 & 0.78 & 5.23\% \\ 
 & 2 & 0.43 & 0.04 & 0.043 & 0.986 & 0.972 & 1.23 & 6.24\% \\ 
\hline 
Cheese & 1 & 0.14 & 0.26 & 0.43 & 0.1557 & 0.319 & 0.82 & 8.41\% \\ 
Box & 2 & 0.12 & 0.12 & 0.45 & 0.1329 & 0.417 & 1.18 & 9.6\% \\ 
\hline 
dice & 1 & 0.05 & 0.05 & 0.479 & 0.603 & 0.606 & 0.56 & 6.77\% \\ 
 & 2 & 0.054 & 0.054 & 0.052 & 0.629 & 0.613 & 0.92 & 6.28\% \\ 
\hline 
Pringles & 1 & 0.012 & 0.012 & 0.168 & 0.355 & 1.28 & 0.95 & 7.48\% \\ 
 Box & 2 & 0.016 & 0.016 & 0.162 & 0.372 & 1.315 & 1.24 & 8.21\% \\ 
\hline
\end{tabular}
\label{tab:singleObject}
\end{scriptsize}
\end{center}
\end{table}

\subsection{Grasping}
\begin{figure}[h!]
\centering
\includegraphics[scale=0.55]{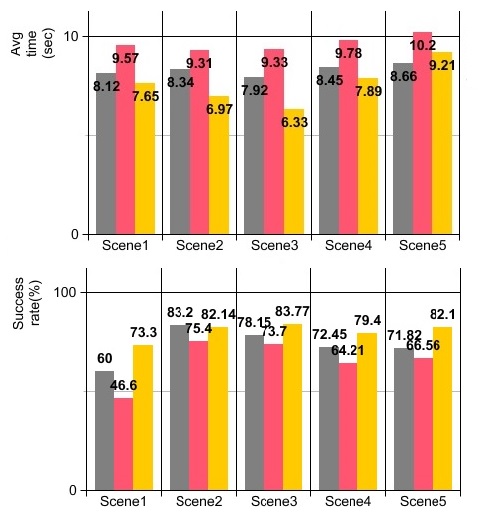}
\caption{Grasping result compared with methods presented in \cite{Platt:2016} and \cite{Vincze:2015}. From left to right: a) avg. time and b)success rate for individual objects. Gray bars are \cite{Platt:2016}, Red bars are \cite{Vincze:2015} and while yellow bars are by Mirroring and SQ fitting  }
\label{fig:combine_graph}
\end{figure}

For grasping objects both in single and dense cluttered scenarios, we compare our results with \cite{Platt:2016} and \cite{Vincze:2015} in Table \ref{tab:singleObject} and Fig. \ref{fig:combine_graph}, as these two methods are well-recognized grasping methods for unknown objects. These two methods have certain advantages over our method as they are learning-based methods. For single objects, the objects are kept in 3 different positions in the environment and the same process of grasping is been iterated 3 times. The average results are presented. For dense cluttered scenes, 5 experiments are performed and for all the experiments and the objects are kept the same for individual experiments. The objects chosen for each experiments are shown in Fig. \ref{fig:fitting}. 

\begin{table}[h]
\label{tab:singlegrasp}
\caption{Object in Isolation}
\begin{center}
\begin{scriptsize}
\begin{tabular}{ |c||c|p{0.5cm}|p{0.5cm}|p{0.5cm}|c|c|}

\hline
Object & Method & Loc. & Loc. & Loc. & Avg & succeess  \\ 
 &  &1 &2 &3 &.Time &  Rate (\%)  \\ 
  
\hline
Ball&   Agile  & 3/3 & 3/3 & 3/3 & 4.24s & 66.67 \\ 
&        Haf   & 3/3 & 1/3 & 2/3 & 6.71s & 66.67\\
&    HyperGrasp & 3/3 & 1/3 & 2/3 & 3.02s & 66.67 \\

\hline
Box&   Agile  & 2/3 & 1/3 & 2/3 & 4.57s & 55.56 \\ 
&        Haf   & 2/3 & 1/3 & 1/3 & 6.32s & 44.45\\
&    HyperGrasp & 2/3 & 1/3 & 2/3 & 3.21s & 55.56 \\
\hline
Toy&   Agile  & 2/3 & 1/3 & 3/3 & 3.78s & 66.67 \\ 
Cylinder& Haf   & 2/3 & 0/3 & 2/3 & 6.45s & 44.45\\
&    HyperGrasp & 3/3 & 2/3 & 3/3 & 3.252s & 88.89 \\
\hline
Screw&   Agile  & 1/3 & 1/3 & 2/3 & 4.11s & 44.45 \\ 
Driver &        Haf   & 2/3 & 0/3 & 1/3 & 6.23s & 33.34\\
&    HyperGrasp & 3/3 & 1/3 & 3/3 & 3.16s & 77.78 \\
\hline
\end{tabular}
\end{scriptsize}
\end{center}
\end{table}

As our system does not rely on iteration for finding grasps to execute, it provides better results in time efficiency than other methods. For single objects, location 2 is the place where every method struggled. The location is chosen deliberately far away from the camera frame, where the segmentation process is prone to failure. As the Agile method is not dependent on segmentation, it provides a better result for only one object (ball). It failed significantly for the rest of the objects. For cluttered scenes in Fig. \ref{fig:fitting} our system consumes much more time for the first and the last scenarios than the other scenarios, as those contains the highest number of objects in a given scene. In experiment 1, though our method calculated all the grasp poses, the dice slipped from the gripper, affecting the accuracy. The vertical placement of the camera where it perceives the environment from the top explains the reason for the significantly poor performance of the Haf grasping method. As for the Agile grasping, the execution system of agile grasping is not based on most vertical grasps, so though in most of the cases, the objects are perceived properly for good grasps, it still failed to execute them.

\section{Conclusions} \label{sec:conclusion}

In this work, we propose a fast antipodal grasping algorithm based on the properties of the SQ, which are fitted to single-view objects. The generation of contact points on the SQ is very fast and does not require a second phase of analysis of the potential contact points to select the optimal candidate. The method has been tested in regular objects and the results show good potential for real time and good grasping success. This method and its implementation is available and updated for open-source community at https://github.com/jontromanab/sq\textunderscore grasp






\section*{ACKNOWLEDGMENTS}

The authors are grateful to Prof. Maya Cakmak for providing us with a PR2 robot we could use for the experiments.



%

\bibliographystyle{ieeetr}
\bibliography{sq}




\end{document}